\pdfoutput=1

\documentclass[11pt]{article}

\usepackage[final]{acl}

\usepackage{times}
\usepackage{latexsym}

\usepackage[T1]{fontenc}

\usepackage[utf8]{inputenc}

\usepackage{microtype}

\usepackage{inconsolata}

\usepackage{graphicx}
\usepackage{amsmath,amssymb}
\usepackage{mathtools}
\usepackage{booktabs}
\usepackage{multirow}
\usepackage{physics}
\usepackage{subcaption}
\usepackage{soul}

\title{Improving Explainability of Sentence-level Metrics via \\ Edit-level Attribution for Grammatical Error Correction}

\author{Takumi Goto, \
  Justin Vasselli, \
  Taro Watanabe \\
  Nara Institute of Science and Technology \\
  \texttt{\{goto.takumi.gv7, vasselli.justin\_ray.vk4, taro\}@is.naist.jp}}

\begin{document}
\maketitle
\begin{abstract}
Various evaluation metrics have been proposed for Grammatical Error Correction (GEC), but many, particularly reference-free metrics, lack explainability.
This lack of explainability hinders researchers from analyzing the strengths and weaknesses of GEC models and limits the ability to provide detailed feedback for users.
To address this issue, we propose attributing sentence-level scores to individual edits, providing insight into how specific corrections contribute to the overall performance.
For the attribution method, we use Shapley values, from cooperative game theory, to compute the contribution of each edit.
Experiments with existing sentence-level metrics demonstrate high consistency across different edit granularities and show approximately 70\% alignment with human evaluations.
In addition, we analyze biases in the metrics based on the attribution results, revealing trends such as the tendency to ignore orthographic edits.
Our implementation is available at \url{https://github.com/naist-nlp/gec-attribute}.
\end{abstract}

\section{Introduction}\label{sec:introduction}

Grammatical error correction (GEC) is the task of automatically correcting grammatical or superficial errors in an input sentence.
Automatic evaluation metrics play a key role in improving GEC performance, but their effectiveness depends on their level of explainability.
For example, metrics that evaluate at the edit level are more explainable than sentence-level metrics, as they allow us to identify which specific edits are effective and which are not, even when a GEC system makes multiple edits.
Such explainable metrics enable researchers to analyze the strengths and weaknesses of GEC models, providing valuable insights into how models can be improved.
Furthermore, in education applications, explainable metrics can provide language learners with detailed feedback on their writing, supporting their learning more effectively.

\begin{figure}[t]
\centering
  \begin{minipage}[b]{0.48\textwidth}
    \centering
    \includegraphics[width=0.99\textwidth]{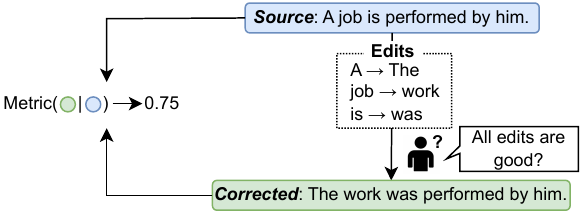}
    \subcaption{The existing metrics are low-explainability.}
    \label{fig:overview-before}
  \end{minipage}
  \begin{minipage}[b]{0.48\textwidth}
    \centering
    \includegraphics[width=0.99\textwidth]{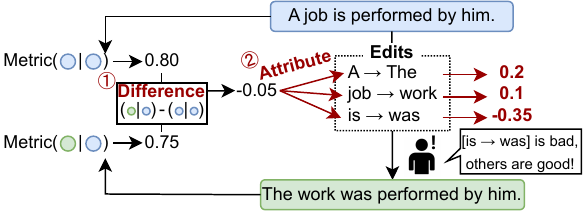}
    \subcaption{Our proposed method improves explainability.}
    \label{fig:overview-after}
  \end{minipage}
  \caption{Overview of the proposed method with an example using three edits. Figure (a) shows the low-explainability of existing metrics that only estimate the sentence-level score, but Figure (b) shows that the edit-level attribution solves this issue.}
    \label{fig:overview}
\end{figure}

In GEC, explainable reference-based metrics, such as ERRANT~\cite{felice-etal-2016-automatic, bryant-etal-2017-automatic} are limited because references cannot account for all valid corrections.
Preparing test data with comprehensive references is often impractical, especially when targeting domains such as medical or academic writing that differ from existing datasets.
To address this issue, reference-free metrics have been proposed to evaluate corrected sentences without relying on references~\cite{choshen-abend-2018-reference, yoshimura-etal-2020-reference, islam-magnani-2021-end, maeda-etal-2022-impara}.
Although these reference-free metrics achieve high correlation with human evaluations, many are designed to assign scores at the sentence level, limiting their explainability on individual edits.
This lack of granularity makes it difficult to analyze how specific edits contribute to the overall sentence score.
For example, as shown in Figure~\ref{fig:overview}, a metric evaluates a corrected sentence created by applying the three edits.
As shown in Figure~\ref{fig:overview-before}, the sentence-level metric assigns an overall score of 0.75, but it does not indicate whether all edits are valid, or if both valid and invalid edits have been applied.

To improve the explainability of metrics with low or no explanation, we propose attributing sentence-level scores to individual edits as illustrated in Figure~\ref{fig:overview-after}.
In the proposed method, the total contribution of all edits is calculated as the difference between the scores of the input sentence and the corrected sentence.
This difference is then attributed to the individual edits.
For example, in Figure~\ref{fig:overview-after}, a difference of -0.05 is distributed among three edits with contributions of 0.2, 0.1, and -0.35.
The attribution results are intrepreted using the sign and magnitude of these scores: the sign indicates whether an edit is the valid or invalid, while the magnitude represents the degree of its influence on the final sentence-level score.
We employ Shapley values~\cite{shapley1953value}  from cooperative game theory to fairly distribute the total score among the edits.
By considering various combinations edits, Shapley values allow us to precisely attribute each edit's contribution to the overall sentence score, offering insights into their individual impact.
Unlike previous feature attribution methods~\cite{lundberg2017unified, sundararajan2017axiomatic}, the proposed method is novel in attributing the difference between the input sentence and the corrected sentence.

In the experiments, we apply the proposed method to two popular reference-free metrics, SOME~\cite{yoshimura-etal-2020-reference} and IMPARA~\cite{maeda-etal-2022-impara}, as well as a fluency metric based on GPT-2~\cite{radford2019language} perplexity.
The results show that the proposed attribution method produces consistent scores across different granularities of edits and that edits with larger absolute attribution scores align more closely with human evaluations.
We introduce Shapley sampling values~\cite{10.5555/1756006.1756007} to mitigate the time-complexity issues of calculating Shapley values.
Additionally, we demonstrate that the proposed method can explain metric decisions at both the sentence and corpus levels, categorized by error types.
These analyses reveal the types of edits that metrics give more weight to, as well as provide insights into the strengths and weaknesses of GEC systems.
 
\section{Background}
\paragraph{Edits in GEC.}
The GEC task aims to correct grammatical errors in a source sentence $S$ and output a corrected sentence $H$.
The differences between $S$ and $H$ are often represented as $N$ edits $\boldsymbol{e} = \{e_i\}_{i=1}^{N}$ to enable evaluation~\cite{dahlmeier-ng-2012-better, bryant-etal-2017-automatic, gong-etal-2022-revisiting, ye-etal-2023-cleme}, ensembling~\cite{tarnavskyi-etal-2022-ensembling}, and post-processing~\cite{sorokin-2022-improved} at the edit level.
These edits can be automatically extracted using edit extraction methods~\cite{felice-etal-2016-automatic, bryant-etal-2017-automatic, belkebir-habash-2021-automatic, korre-etal-2021-elerrant, uz-eryigit-2023-towards}.
Each edit typically includes a word-level span in $S$ and its corresponding correction, although it may also include an error type~\cite{bryant-etal-2017-automatic}.
The error type categorizes each edit, indicating the part-of-speech or grammatical aspect it relates to, which helps to analyze the strengths and weaknesses of the GEC systems.

\paragraph{Sentence-level Metrics.}
A sentence-level metric $M$ computes the score of the corrected sentence given the source sentence, denoted as $M(H | S) \in \mathbb{R}$.
The source sentence is used to assess meaning preservation, as GEC requires correcting errors while maintaining the original meaning of the source sentence.
This formulation has been adopted by several reference-free metrics~\cite{yoshimura-etal-2020-reference, islam-magnani-2021-end, maeda-etal-2022-impara, kobayashi-etal-2024-large}.
Sentence-level metrics aim to rank GEC systems in alignment with humans judgments, as evidenced by the fact that the meta-evaluation is performed using the correlation between metric-generated rankings or scores and those of humans.
However, these metrics are limited to sentence-level scoring and cannot explain how individual edits contribute to the final score.

\section{Method}\label{sec:method}
Our attribution method assumes that the overall contribution of edits is the difference in scores before and after correction.
We distribute the difference $\Delta M(H | S) = M(H | S) - M(S | S)$ across each edit $\boldsymbol{e} = \{e_i\}_{i=1}^{N}$, where $M(S | S)$ is the score of the source sentence treated as its own corrected sentence.

The goal of our attribution method is to compute the contribution for each edit denoted as $\{\phi_i (M) \in \mathbb{R}\}_{i=1}^{N}$, so that the following equation is satisfied:
\begin{equation}\label{eq:objective}
    \Delta M(H | S) = \sum_{i=1}^{N} \phi_i (M).
\end{equation}
We refer to $\phi_i (M)$ as \textit{attribution scores}.
A positive score ($\phi_i (M) > 0$) indicates an edit that improves the metric $M(\cdot)$, while a negative score ($\phi_i (M) < 0$) indicates an edit that worsens it.
The absolute value $|\phi_i (M)|$ represents the degree of the edit's impact.

\paragraph{Shapley.}
For the attribution method, we introduce Shapley values~\cite{shapley1953value} from cooperative game theory.
In cooperative game theory, multiple players work together towards a common goal and share the total benefit based on their contributions.
Shapley values distribute this benefit among players fairly, ensuring that those players who contributes more receive a larger share.
For our purpose, we regard $\Delta M(H | S)$ as the total benefit, edits $\boldsymbol{e}$ as the players, and $\phi_i (M)$ as the Shapley values.
The Shapley value $\phi_i (M)$ for a given metric $M(\cdot)$ is calculated as follows:
\begin{equation}
\begin{split}
\label{eq:shapley}
    \phi_i (M) = & \sum_{\boldsymbol{e}' \subseteq \boldsymbol{e} \setminus \{e_i\}}\frac{|\boldsymbol{e}'|! (N-|\boldsymbol{e}'|-1)!}{N!} \\
    & (\Delta M(S_{\boldsymbol{e}' \cup \{e_i\}} | S) -  \Delta M(S_{\boldsymbol{e}'} | S)),
\end{split}
\end{equation}
where $S_{\boldsymbol{e}}$ denotes the source sentence after applying the edit set $\boldsymbol{e}$.
Equation~\ref{eq:shapley} calculates the weighted sum of the differences in evaluation scores when including and excluding the edit $e_i$.
For example, using Figure~\ref{fig:overview} with $\boldsymbol{e} = \{e_1, e_2, e_3\} = \{[\mathrm{A \rightarrow The}], [\mathrm{job \rightarrow work}], [\mathrm{is \rightarrow was}]\}$, one of the terms in the calculation for $\phi_1 (M)$ with $\boldsymbol{e}' = \{e_2\}$ is
\begin{equation}\label{eq:example-shapley}
\begin{split}
    \frac{1}{6}&\qty( \Delta M(S_{\{e_1, e_2\}} | S) - \Delta M(S_{\{ e_2 \}} | S)) \\
    &= \frac{1}{6} ( \Delta M(\text{\textbf{The} \ul{work} \ul{is} performed by him.} | S)
    \\&- \Delta M(\text{\textbf{A} \ul{work} \ul{is} performed by him.} | S)).
\end{split}
\end{equation}
Here, bold words indicate the edit being attributed, and underlined words show other edits.
The terms for $\boldsymbol{e}' = \{\phi \}$, $\{e_3\}$, and $\{e_2, e_3\}$ are computed in a similar way.
Shapley values consider various combinations of edits, ensuring accurately attribution of the $i$-th edit's contribution.
By design, Shapley values naturally satisfy Equation~\ref{eq:objective} due to their \textit{effectiveness}~\cite{shapley1953value}.
However, the computational complexity is $\mathcal{O}(2^N)$.

\paragraph{Shapley Sampling Values.}
To improve computational efficiency, we introduce Shapley sampling values~\cite{10.5555/1756006.1756007}, an approximation of Shapley values. Equation~\ref{eq:shapley} can be rewritten as:
\begin{equation}\label{eq:shaply-in-permutation}
\begin{split}
    \phi_i &(M) = \frac{1}{N!} \sum_{\boldsymbol{o} \in \pi (\boldsymbol{e})} \\
    &(\Delta M(S, S_{\mathrm{Pre}^i(\boldsymbol{o}) \cup \{e_i\}})) - \Delta M(S, S_{\mathrm{Pre}^i(\boldsymbol{o})})) 
\end{split}
\end{equation}
where $\pi(\boldsymbol{e})$ is the set of all possible orders of edits, and $\mathrm{Pre}^i(\boldsymbol{o})$ is the set of edits preceding $e_i$ in permutation $\boldsymbol{o}$.
In the example from Equation~\ref{eq:example-shapley}, $\mathrm{Pre}^1(\boldsymbol{o}) = \{\phi\}$ when $\boldsymbol{o} = [e_1, e_2, e_3]$, and $\mathrm{Pre}^1(\boldsymbol{o}) = \{e_2, e_3\} = \{[\mathrm{job \rightarrow work}], [\mathrm{is \rightarrow was}]\}$ when $\boldsymbol{o} = [e_3, e_2, e_1]$.
To approximate Shapley values, we uniformly sample $T$ permulations without replacement from $\pi (\boldsymbol{e})$, denoted as $\overset{\sim}{{\pi (\boldsymbol{e})}} = \{\boldsymbol{o}_1, \dots, \boldsymbol{o}_T\}$.
Shapley sampling values are then calculated using $\overset{\sim}{{\pi (\boldsymbol{e})}}$ instead of $\pi (\boldsymbol{e})$ in Equation~\ref{eq:shaply-in-permutation}.
This approximation reduces the computational cost from $\mathcal{O}(2^N)$ to $\mathcal{O}(TN)$.

\paragraph{Normalized Shapley Values}
The calculated attribution scores are not directly comparable across different sentence-level scores.
For instance, an attribution score of 0.2 has a different relative impact when distributing a sentence-level score of 1.0 versus -0.05.
To enable meaningful comparison, we apply L1 normalization to the attribution scores:
\begin{equation}
    \phi_i^{\text{norm}}(M) = \frac{\phi_{i}(M)}{\sum_{i=1}^{N} |\phi_i (M)|}.
\end{equation}
This normalization, applied as a post-processing step, adjusts only the magnitude of the scores while preserving their original signs.
Since the normalized scores represent the ratio of each edit's contribution, they are assumed to be comparable even when the sentence-level scores differ.

\section{Evaluation of Attribution}\label{sec:experiments}
We evaluate the proposed attribution method from two perspectives: faithfulness and explainability~\cite{wang2024gradientbasedfeatureattribution}.
Faithfulness measures how well the attribution results reflect the model's internal decision, while explainability assesses the extent to which the results are understandable to humans.
To demonstrate the effectiveness of the proposed method across various domains, we conduct experiments using diverse datasets, GEC systems, and metrics.

\subsection{Experimental Settings}

\subsubsection{Datasets}\label{sssec:dataset}
We use the CoNLL-2014 test set~\cite{ng-etal-2014-conll} and the JFLEG validation set~\cite{heilman-etal-2014-predicting, napoles-etal-2017-jfleg}.
CoNLL-2014 is a benchmark for minimal edits, focusing on correcting errors while preserving the original structure of the input as much as possible.
In contrast, JFLEG is a benchmark for fluency edits, allowing more extensive rewrites to produce fluent and natural sentences.

\subsubsection{GEC Systems}\label{sssec:gec-system}
\begin{figure}
    \centering
    \includegraphics[width=0.45\textwidth]{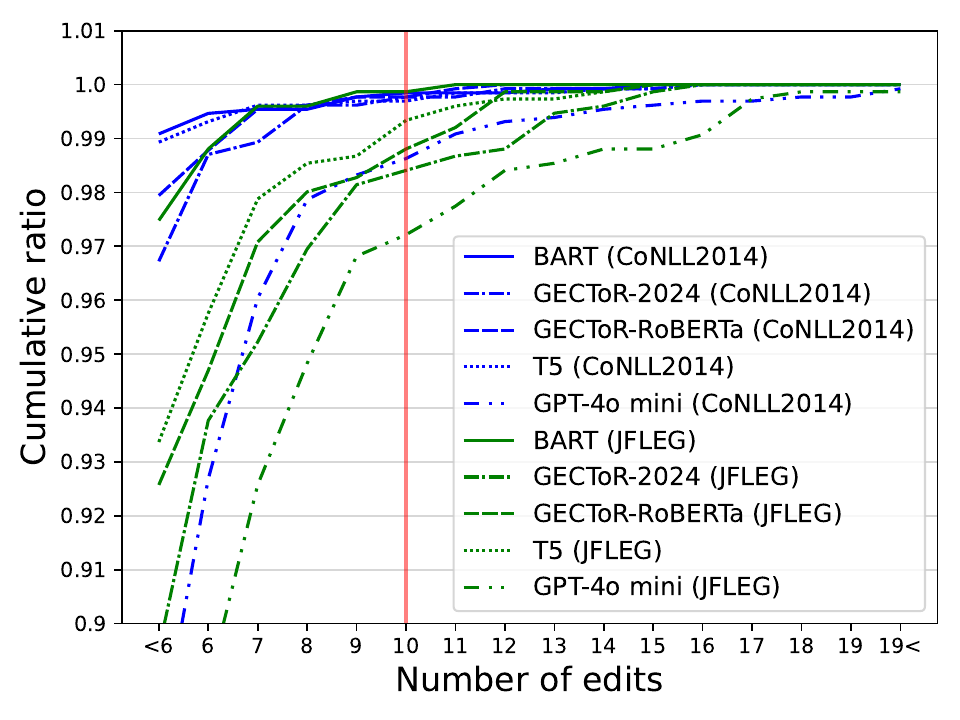}
    \caption{Cumulative sentences ratio regarding the number of edits. The red line indicates the position where the number of edits is 10.}
    \label{fig:num_edits}
    \vspace{-0.5cm}
\end{figure}

We evaluate our attribution method on various GEC systems, including two tagging-based models (the official RoBERTa-based GECToR~\cite{omelianchuk-etal-2020-gector} and GECToR-2024~\cite{omelianchuk-etal-2024-pillars}),
two encoder-decoder models (BART~\cite{lewis-etal-2020-bart} and T5~\cite{rothe-etal-2021-simple}), and a causal language model (GTP-4o mini).
This allows us to assess the explainability of attributions scores across different GEC architectures.
For GPT-4o mini, we used a two-shot setting following ~\citet{coyne2023analyzingperformancegpt35gpt4}, with examples randomly sampled once from the W\&I+LOCNESS validation set~\cite{yannakoudakis2018developing} and used for all input sentences.
Note that we use only the corrected sentences containing 10 or fewer edits ($N \leq 10$) due to the computational complexity of Shapley values.
According to Figure~\ref{fig:num_edits}, which shows the cumulative sentence ratio by the number of edits, our experiments cover at least more than 97\% of the sentences in both datasets.

\begin{figure*}[t]
\centering
  \begin{minipage}[b]{0.95\textwidth}
    \centering
    \includegraphics[width=0.99\textwidth]{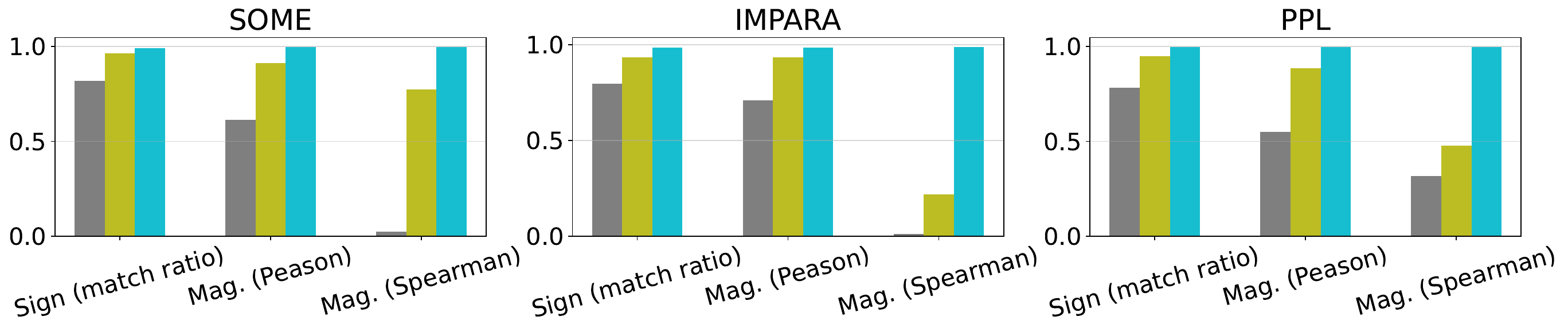}
    \subcaption{CoNLL-2014 results.}
  \end{minipage}
  \begin{minipage}[b]{0.95\textwidth}
    \centering
    \includegraphics[width=0.99\textwidth]{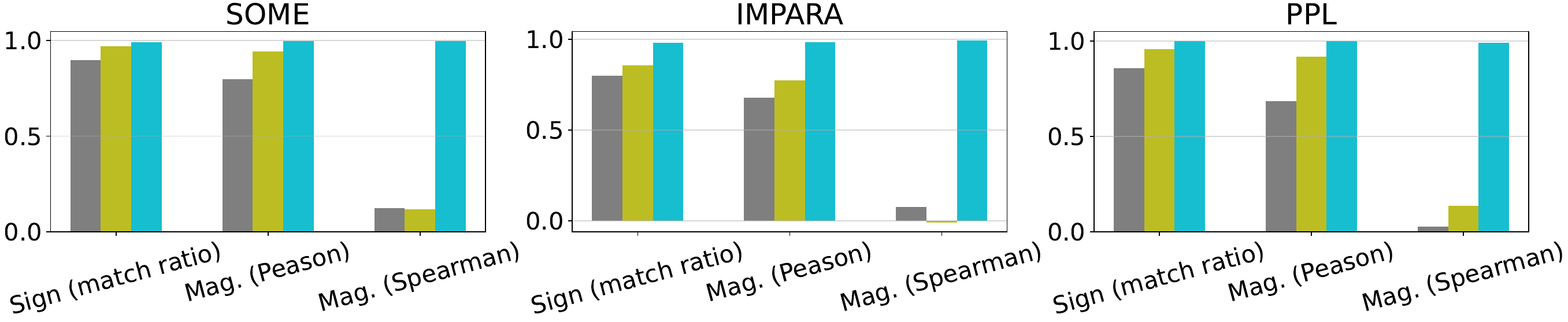}
    \subcaption{JFLEG results.}
    \includegraphics[width=0.40\textwidth]{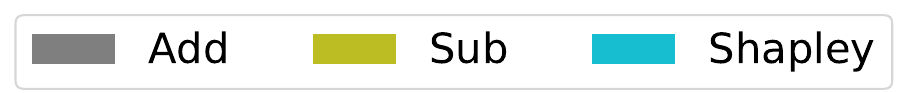}
  \end{minipage}
  \caption{The results of consistency-based evaluation. Each row shows the different datasets and each column shows different metrics. ``Mag.'' means the magnitude. Colors show the attribution scores.}
  \label{fig:consis-base-assess}
\end{figure*}

\subsubsection{Reference-free Metrics}

\paragraph{SOME \cite{yoshimura-etal-2020-reference}}
trains a BERT-based regression model optimized directly on human evaluation results.
We used the official pretrained model weights\footnote{\url{https://github.com/kokeman/SOME}} and used the default coefficients for the weighted average of grammaticality, fluency, and meaning preservation scores, from the official script\footnote{0.55*grammaticality + 0.43 * fluency + 0.02 * meaning preservation.}.

\paragraph{IMPARA \cite{maeda-etal-2022-impara}}
estimates evaluation scores through similarity estimation and quality estimation.
We use BERT (\texttt{bert-base-cased}) as the similarity estimator and train our own model for the quality estimator, as the official pre-trained weights are not available.
Our quality estimator was trained following the same settings described in \citet{maeda-etal-2022-impara}, achieving a correlation with the human ranking comparable to their reported results.

\paragraph{GPT-2 Perplexity (PPL).}
Our proposed method can be applied to metrics that evaluate only the quality of the corrected sentence\footnote{In this case, the sentence-level score is $\Delta M(S, H) = M(H) - M(S)$}.
To test this, we use GPT-2~\cite{radford2019language} perplexity, with negative perplexity scores to ensure that higher values correspond to better quality.

\subsection{Baseline Attribution Methods}\label{subsec:baselines}
To evaluate the effectiveness of Shapley values, we employ simpler variants, i.e., \texttt{ADD} and \texttt{Sub}, as baseline attribution methods.

\paragraph{Add.}
This method observes the change in the score when each edit is applied individually to the source sentence.
An edit that increases the score is considered valid for the metric.
This approach corresponds to using only $\boldsymbol{e}' = \{\phi\}$ in Equation~\ref{eq:shapley}, with the attribution scores normalized by $\frac{\Delta M(H | S)}{\sum_{i=1}^{N} \phi_i(M)}$ so that it satisfies Equation~\ref{eq:objective}.

\paragraph{Sub.}
This method observes the change in the score when each edit is removed individually from the corrected sentence.
An edit that decreases the score upon removal is considered valid for the metric.
This approach corresponds to using only $\boldsymbol{e}' = \boldsymbol{e} \setminus \{e_i\}$ in Equation~\ref{eq:shapley}, with the attribution scores normalized by $\frac{\Delta M(H | S)}{\sum_{i=1}^{N} \phi_i(M)}$ so that it satisfies Equation~\ref{eq:objective}.

\subsection{Consistency Evaluation}\label{subsec:consis-eval}

To evaluate faithfulness, we test how well the attribution scores represent the judgments of the metrics through consistency evaluation.
Specifically, we first calculate the attribution scores for individual edits and then group edits with the same sign, treating them as a single edit.
Next, we calculate the attribution score for the grouped edits. 
We hypothesize that the attribution score for a grouped edit should equal the sum of the individual attribution scores of the edits comprising the group.
If this condition holds, the attribution method consistently calculates the contributions of edits, making its results reliable for practical use.
We use an agreement ration to measure the consistency of the signs and use Pearson and Spearman correlations to assess the consistency of the magnitudes.

For example, in Figure~\ref{fig:overview}, we group two positivity-attributed edits, [\textit{A → The}] and [\textit{job → work}], into a single edit and compute attribution scores for the grouped edit and the remaining edit, [\textit{is → was}].
Ideally, the attribution score for the grouped edit should be $0.2 + 0.1 = 0.3$, which can be verified by sign agreement and closeness to 0.3.

Figure~\ref{fig:consis-base-assess} presents the results for each metrics.
Our proposed Shapley method shows higher consistency than the baseline attribution methods across various domains and metrics.
While the Sub metric also demonstrates high consistency, its Spearman's rank correlation occasionally drops for certain metrics, such as IMPARA.
Low rank correlation can misrepresent the relative importance of edits, posing a serious issue for explainability.
These results suggest that the attribution method is reliable across different edit granularities, such as edits extracted by ERRANT~\cite{felice-etal-2016-automatic, bryant-etal-2017-automatic} or chunks created by merging multiple edits~\cite{ye-etal-2023-cleme}.
This flexibility enables a wide range of applications for the proposed method.

\subsection{Human Evaluation}\label{subsec:ref-eval}

To evaluate explainability, we assess the agreement between attribution scores and human evaluation results using references.
Ideally, a positively attributed edit should align with a correct edit in the reference-based evaluation, while a negativity attributed edit should correspond to an incorrect one.
Furthermore, edits with larger absolute attribution scores are expected to show higher agreement with human evaluations.

In this experiment, we annotate two types of labels for each edit: one based on the sign of the attribution score and another based on reference-based evaluation.
We then calculate the matching ratio between these labels at the corpus level.
For the evaluation, we use the two official references for CoNLL-2014, and four official references for JFLEG validation set.
The assessment is performed on mixed outputs from five GEC systems.
To ensure the analysis focuses on meaningful cases, we include only sentences with two or more edits.
When assigning labels for reference-based evaluation with multiple references, we select the reference that results in the highest agreement with the attribution scores.
To further examine the relationship between the magnitude of attribution scores and agreement rates, we follow standard attribution evaluation practices~\cite{petsiuk2018rise, fong2017interpretable} by applying a threshold to the absolute values of the scores.
We use only edits with normalized absolute attribution scores below the threshold for accuracy calculations.
The threshold starts at 0.1 and increases in steps of 0.1 until it reaches 1.0, where all edits are included.

\begin{figure}[t]
\centering
  \begin{minipage}[b]{0.49\textwidth}
    \centering
    \includegraphics[width=0.80\textwidth]{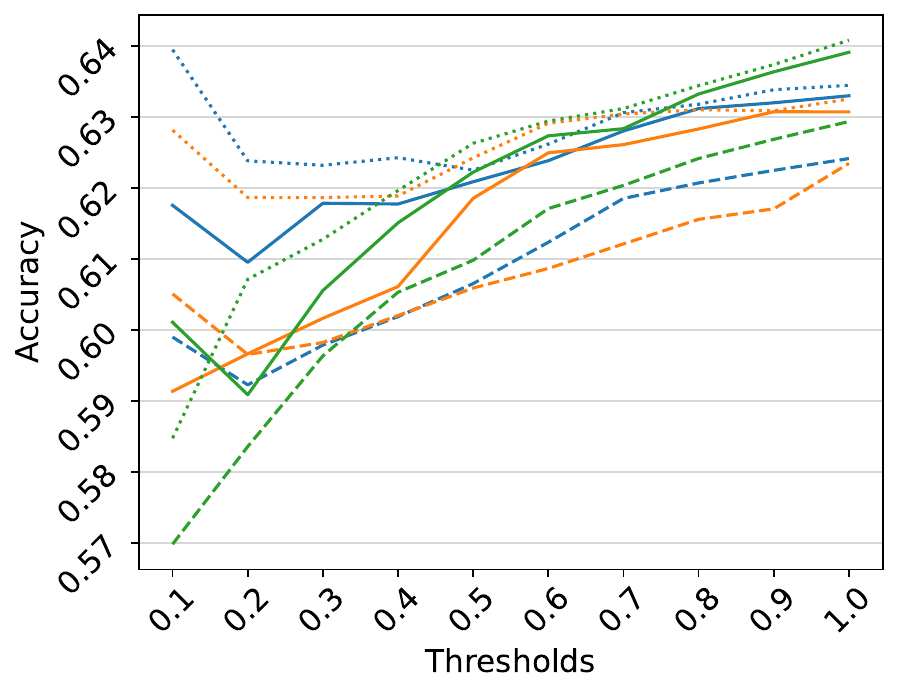}
    \subcaption{CoNLL-2014 results.}
    \label{fig:human-eval-conll14}
  \end{minipage}
  \begin{minipage}[b]{0.49\textwidth}
    \centering
    \includegraphics[width=0.80\textwidth]{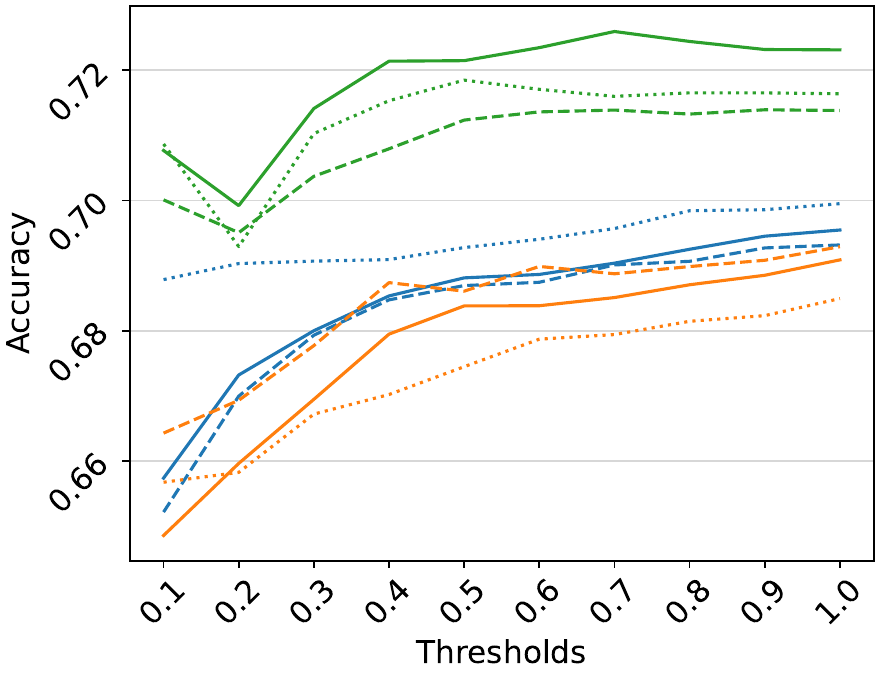}
    \subcaption{JFLEG results.}
    \includegraphics[width=\textwidth]{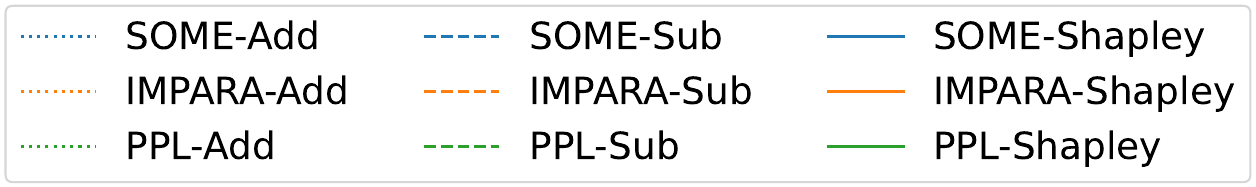}
    \label{fig:human-eval-jfleg}
  \end{minipage}
  \caption{Human evaluation results for CoNLL-2014 and JFLEG. Colors indicate metrics and line styles indicate attribution methods.}
  \label{fig:human-evaluation}
\end{figure}

Figure~\ref{fig:human-evaluation} presents the results for the CoNLL-2014 and JFLEG datasets.
Overall, the results show that including edits with larger absolute attribution scores improves the agreement with human evaluation, indicating that the magnitude of these scores is meaningful.
When comparing attribution methods, Shapley rarely achieves the worst agreement.
For instance, in JFLEG, the SOME metric shows the order Add > Shapley > Sub, while the IMPARA metric shows Sub > Shapley > Add.
Either Add or Sub often results in the worst agreement, whereas Shapley demonstrates more stable performance across different metrics and domains.

When comparing metrics,
particularly in the results for JFLEG (Figure~\ref{fig:human-eval-jfleg}), the agreement rates consistently rank in the order of PPL, SOME, and IMPARA.
This trend may reflect the characteristics of these reference-free metrics in relation to reference-based evaluation.
In fact, when we compute the correlation with ERRANT~\footnote{We use ERRANT as a representative edit-based and reference-based metric.} using standard sentence-level meta-evaluation~\cite{kobayashi2024revisiting}, the rankings follow the same order: of PPL (0.550), SOME (0.529), and IMPARA (0.516), with Kendall rank correlation coefficients of 0.100, 0.058, and 0.033, respectively.
These results suggest that metrics more closely aligned with reference-based evaluation can be attributed more accurately, improving the reliability of our attribution method.
On the other hand, for CoNLL-2014, the sentence-level correlation shows the order of PPL (0.522), IMPARA (0.479), and SOME (0.477). However, the agreement in Figure~\ref{fig:human-eval-conll14} does not follow this trend.
This indicates that the proposed method aligns well with human judgement in case of fluency edits.
Conversely, minimal edits may require further studies, but primarily depend on the development of better reference-free metrics.

\begin{table*}[t]
    \centering
    \small
    \scalebox{1}{
    \begin{tabular}{l|c|cccccccccc}
    \toprule

Original ($S$) & - & Further more &  & by & these & evidence &  & u & will agree & \\
Correction ($H$) & - & Further more & ,  & with & this & evidence & , & you & will agree  & . \\
\midrule
\midrule
Metrics ($M$) & $\Delta M(\cdot)$ & \multicolumn{9}{c}{Shapley values $\phi_i (M)$} \\
\midrule
SOME & 0.298 & - & 0.068 & 0.064 & 0.033 &    -     & 0.038 & 0.066 &     -      & 0.030 \\
IMPARA & -0.027 & - & 0.068 & 0.029 & 0.124 &    -     & 0.145 & -0.361 &     -      & -0.033  \\
PPL & 1266.3 & - & 250.7 & 103.8 & 216.0 &    -     & 67.4 & 366.6 &     -      & 261.5  \\
\midrule
& & \multicolumn{9}{c}{Normalized Shapley values} \\
\midrule
SOME &  & - & 0.229 & 0.215 & 0.111 &    -     & 0.126 & 0.220 &     -      & 0.099 \\
IMPARA &  & - & 0.090 & 0.039 & 0.163 &    -     & 0.191 & -0.475 &     -      & -0.043 \\
PPL &  & - & 0.198  &  0.082  &  0.171  &    -     & 0.053  &  0.290  &     -      &  0.207 \\
    \bottomrule
    \end{tabular}
    }
    \caption{An example of the proposed method's results using actual sentence.}
    \label{tab:case-study}
\end{table*}

\subsection{Efficiency of Shapley Values}\label{subsec:efficiency}
\begin{figure}[t]
    \centering
    \includegraphics[width=0.45\textwidth]{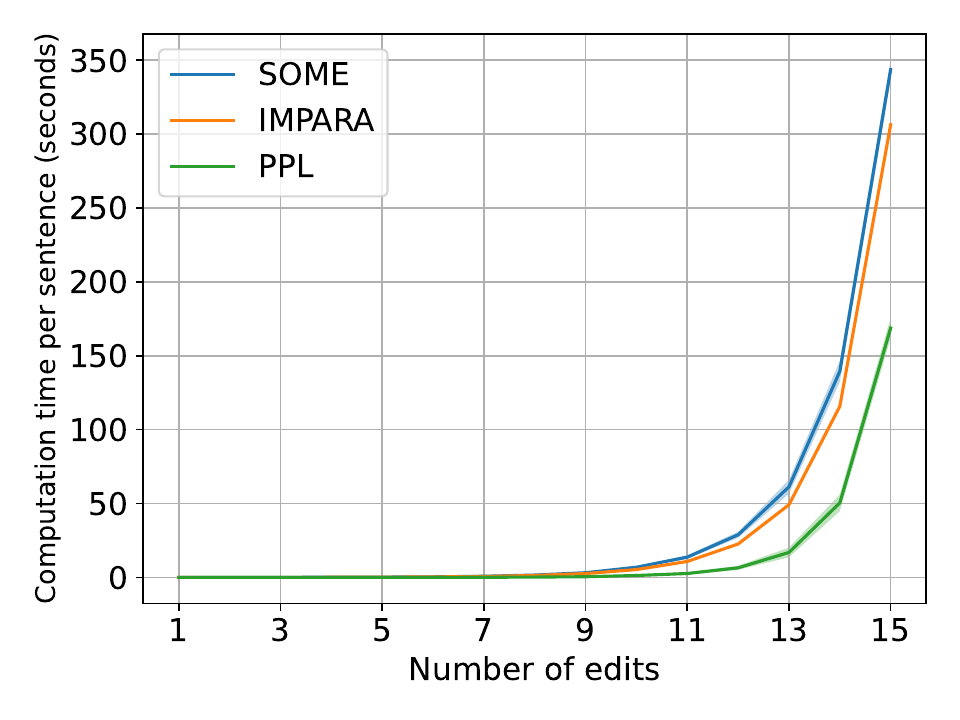}
    \caption{The relationship between the number of edits and computation time per sentence. The solid lines are average time and ranges are standard deviation.}
    \label{fig:time-comp}
    \vspace{-0.5cm}
\end{figure}

One limitation of Shapley values is their high computational cost.
Figure~\ref{fig:time-comp} shows the relation between the number of edits and the computation time per sentence in seconds on a single RTX 3090.
The computation time increases rapidly when the number of edits exceeds 11.
For this reason, we assume that sentences with more than 11 edits are impractical to attribute within a reasonable time.
According to Figure~\ref{fig:num_edits}, the affects approximately 3\% of the sentences in GEC output.
Similarly, tasks involving a higher number of edits, such as text simplification, could face even greater challenges.

As discussed in Section~\ref{sec:method}, we address this issue by employing Shapley sampling values and evaluate their ability to approximate exact Shapley values by measuring the average absolute differences between them.
For system-independent experiments, we use a dataset combining all GEC model corrections on the JFLEG validation set.
We set $T = 64$ and restrict sentences to $10 \leq N \leq 15$
\footnote{When $T=64$ and $10 \leq N$, the computation cost of Shapley sampling values is consistently lower than that of exact Shapley values, as $2^x > 64x$ holds for $x > 9.20\dots$.}.

Table~\ref{tab:sampling-stat} reports the errors and computation times for each metric.
With Shapley sampling values, the computation time per sentence can be reduced to as little as one second.
To assess the impact of errors, we also show the distribution of absolute original Shapley values. 
While SOME and PPL show errors below the average, IMPARA exhibits higher errors.
This discrepancy with IMPARA can lead to misinterpretations of attribution scores.
For example, the frequency of changes in the relative contributions of different edits is likely to increase, undermining reliability.
IMPARA's higher error rate may be due to its smaller variance in evaluated values, making it less effective at quantifying impact with a limited number of calculations.

\begin{table}[]
    \centering
    \small
    \begin{tabular}{l|cc|c}
    \toprule
        Metric & Error & Time & Shapley values dist. \\
        \midrule
        SOME & 0.014 & 1.00 & 0.019 ± 0.020 \\
        IMPARA & 0.066 & 0.99 & 0.052 ± 0.071  \\
        PPL & 17.515 & 0.20 & 34.549 ± 59.472  \\
        \bottomrule
    \end{tabular}
    \caption{The average error and average computation time (seconds) when using Shapley sampling values. It also shows the distribution of the absolute original Shapley values (the average ± the standard deviation).}
    \label{tab:sampling-stat}
\end{table}

\section{Applications of Attribution Scores}
We demonstrate practical applications of attribution scores for users.
All results in this section are based on Shapley values for the attribution method.

\subsection{Case Study}\label{subsec:case-study}

Attribution scores can be used to identify which edits improve or worsen the sentence-level score.
Table~\ref{tab:case-study} provides an example, showing attribution scores and their normalized version.
The original sentence and its corrections are chunked according to edit spans, omitting scores for non-edited chunks which are all zeros.
One observation is that the sentence-level score of IMPARA declines primarily due to the edit [\textit{u → you}], as identified by the attribution score.
In contrast, SOME and PPL prefer this edit.
This analysis demonstrates how attribution scores can reveal weaknesses in metrics as seen in Table~\ref{tab:case-study}.

Normalized Shapley values enable comparison of attribution scores across metrics.
For example, while SOME and IMPARA assign the same Shapley value to the edit [\textit{$\phi$ → ,}], their normalized scores reveal differing impacts.
This feature is particularly useful for comparing metrics with different value ranges, such as SOME and PPL.

\begin{figure}[t]
    \centering
    \includegraphics[width=0.48\textwidth]{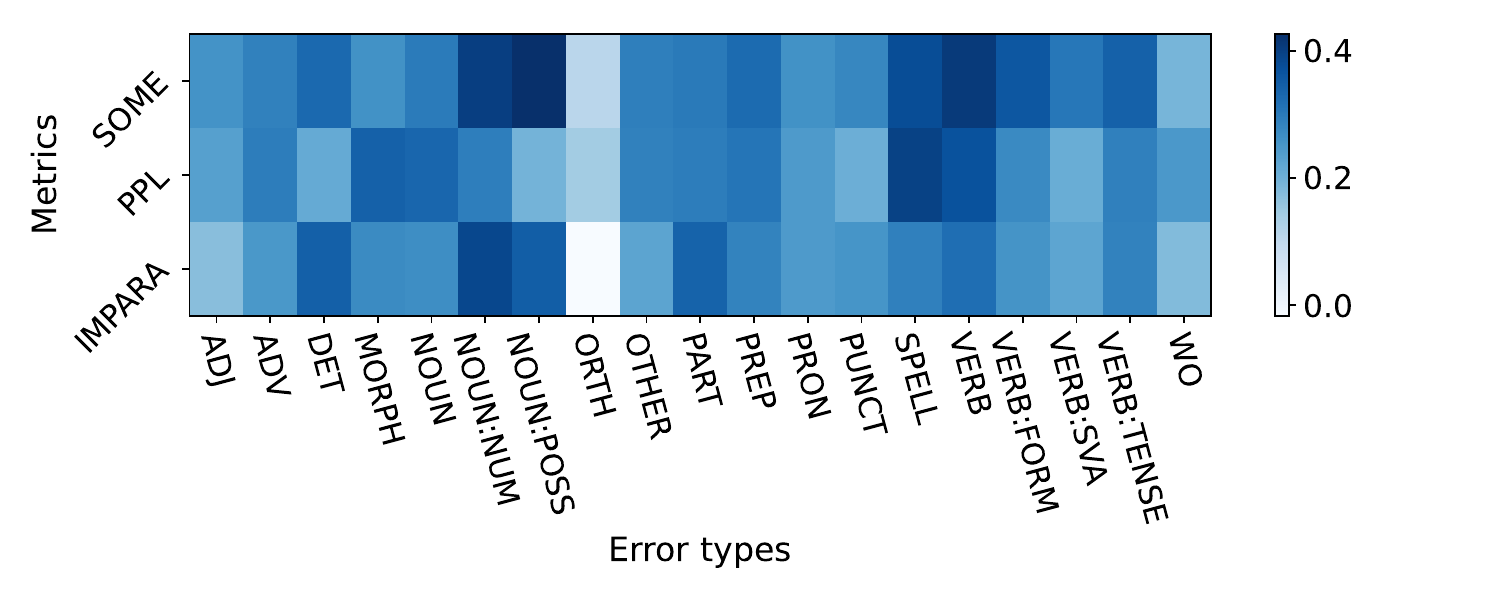}
    \caption{The heatmap indicating the average of normalized Shapley values per error type. The deeper color indicates higher values.}
    \label{fig:error-type-bias}
\end{figure}

However, the metrics themselves may exhibit biases that affect attribution scores.
To investigate these biases, we calculate the average normalized Shapley values for each error type~\cite{bryant-etal-2017-automatic}.
As in Section~\ref{subsec:efficiency}, we combine the corrected sentences from five GEC systems for the JFLEG validation set to mitigate biases specific to individual GEC models.
Figure~\ref{fig:error-type-bias} shows a heatmap of average normalized attribution scores for error types with a frequency greater than 30.
The results indicate that different metrics emphasize different error types.
For instance, orthography (\texttt{ORTH})
edits, such as case changes and whitespace adjustments, tend to be downplayed.
Metric biases must be considered when interpreting attribution scores.
It is important to not that the attribution scores reflect the internal decisions of the metric and may not align with the true correctness of edits.
We leave addressing these biases to future work.

\subsection{Precision per Error Type}\label{subsec:precision}
While the case study focused on local, sentence-level evaluation, the proposed method can be extended to corpus-level analysis.
Typically, metrics with low explainability provide only a single numerical score at the corpus level. 
By applying the proposed method, we can decompose this score is into performance across different error types.
Specifically, we treat edits with positive attribution scores as True Positives, and those with negative attribution scores as False Positives, enabling the calculation of precision for each error type.
To handle attribution scores across multiple sentences, we use normalized Shapley values:
\begin{equation}
    \mathrm{Precision} = \frac{
        \phi_+^{\text{norm}}(M)
    }{
        \phi_+^{\text{norm}}(M) + |\phi_-^{\text{norm}}(M)|
    },
\end{equation}
where $\phi_+^{\text{norm}}(M)$ and $\phi_-^{\text{norm}}(M)$ represent the sum of positive and negative normalized attribution scores at the corpus-level, respectively.
This is similar to PT-M2~\cite{gong-etal-2022-revisiting} which proposed an edit-level weighted evaluation. However, our method is designed to enhance the corpus-level explainability of metrics rather than to improve agreement with human evaluations.%

\begin{figure}[t]
    \centering
    \includegraphics[width=0.48\textwidth]{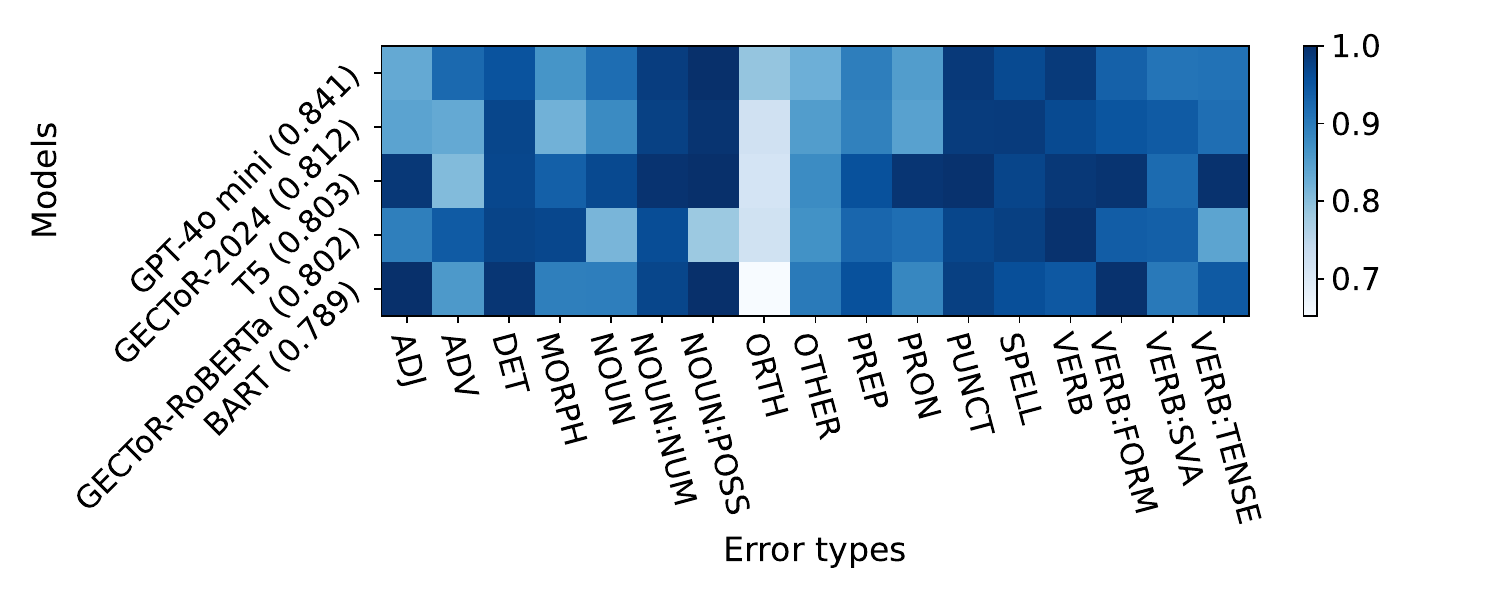}
    \caption{The heatmap indicating the precision for each GEC systems. We used JFLEG validation set as a dataset and SOME as a metric.}
    \label{fig:precision}
\end{figure}

Figure~\ref{fig:precision} shows the precision for each error type using the JFLEG validation set and SOME as the evaluation metric.
The parentheses in the y-axis labels indicate the corpus-level scores, with each row of the heatmap explaining these score in terms of error types.
The results reveal that better edits in adverbs (ADV) or orthography (ORTH) contribute most to the highest corpus-level score achieved by GPT-4o mini.
On the other hand, despite achieving the highest corpus-level score among the five systems, GPT-4o mini's precisions are not particularly high.
Notably, T5 appears to perform better in terms of precision, as indicated by more dark-colored cells.
This discrepancy may stem from an overcorrection issue, leading to a low-precision, high-recall trend in performance~\cite{fang2023chatgpthighlyfluentgrammatical, omelianchuk-etal-2024-pillars}.
While this trend is intuitive because the valid edits in the reference-based evaluation are limited to the references, we also observe a similar trend even for reference-free evaluation metrics.

\section{Conclusion}
This paper proposes a method to improve the explainability of existing low-explainable GEC metrics by attributing sentence-level scores to individual edits.
Specifically, we employed Shapley values to perform attribution while accounting for various contexts in which edits are applied.
Quantitative evaluations showed that the attribution scores align with metric's judgement achieve approximately 70\% agreement with human evaluations.
Additionally, we demonstrated how attribution scores can be used at both the sentence and corpus levels. Finally, we discussed the biases of existing metrics.

\section*{Limitations}
\paragraph{Treating False Negative Corrections.}
As mentioned in Section~\ref{subsec:precision}, the proposed method is limited to analyzing corrections made by the GEC system, i.e. True Positives (TP) and False Positives (FP), and does not address False Negatives (FN).
While we assume that the effect of FN corrections is canceled out by $\Delta M(H | S) = M(H | S) - M(S | S)$, it may still affect the computation of attribution scores.
A more detailed investigation into this issue is left for future work.

\paragraph{Treating dependent edits}
Edits might exhibit dependencies.
For example, the correction [\textit{model 's prediction} -> \textit{prediction of the model}] can be split into two dependent edits: [\textit{model 's} -> $\phi$] and [$\phi$ -> \text{of the model}].
While analyzing these edits together may better capture their contribution, the proposed method evaluates each edit independently.
We assume that Shapley values partially capture such dependent edits by considering various patterns of applying edits.
However, understanding dependencies fully requires error correction data annotated for edit dependencies or tools to automatically identify them.
Developing such resources is left as future work.

\paragraph{Real Human Evaluation}
Unlike Section~\ref{subsec:ref-eval}, which uses a reference-based evaluation framework, we could also conduct direct human evaluation.
However, we prioritize reference-based evaluation for its scalability when applying the method to new metrics or datasets.
It is important to note that the primary goal of this study is not to derive attribution scores that align with human evaluation, but to explain the decision-making process of metrics at the edit level.%
Verifying alignment with human evaluations is a secondary finding.
If the goal were to achieve consistency with human evaluation, training a dedicated model would be a more appropriate approach.

\paragraph{Rectifying Metric Biases }
The case study results (Section~\ref{subsec:case-study}) revealed that metrics exhibit biases towards specific error types.
While one could attempt to mitigate such biases, we believe that sentence-level metrics benefit from implicitly weighting edits, making these biases beneficial for evaluation.
However, biases related to social factors such as gender or nationality, should be addressed.
A deeper investigation into metric biases is beyond the scope of this work, but remains an important area for future research.
Our work provides a strong foundation for exploring these biases

\section*{Acknowledgments}

\bibliography{custom}

\appendix

\end{document}